%% file: compositional_navigation_main.tex
\documentclass{article}
\usepackage{microtype}
\usepackage{graphicx}
\usepackage{subfigure}
\usepackage{booktabs} 

\usepackage{hyperref}

\usepackage{stmaryrd}
\usepackage{sidecap} 
\usepackage{footmisc}

\usepackage[accepted]{icml2018}

\icmltitlerunning{Generalization without Systematicity}

\begin{document}

\twocolumn[
\icmltitle{Generalization without Systematicity:\\On the Compositional Skills of Sequence-to-Sequence Recurrent Networks}



\icmlsetsymbol{equal}{*}

\begin{icmlauthorlist}
\icmlauthor{Brenden Lake}{nyu,fair}
\icmlauthor{Marco Baroni}{fair}
\end{icmlauthorlist}

\icmlaffiliation{nyu}{Dept.~of Psychology and Center for Data Science, New York University}
\icmlaffiliation{fair}{Facebook Artificial Intelligence Research}

\icmlcorrespondingauthor{Brenden Lake}{brenden@nyu.edu}
\icmlcorrespondingauthor{Marco Baroni}{mbaroni@fb.com}

\icmlkeywords{Compositionality, Recurrent neural networks}

\vskip 0.3in
]



\printAffiliationsAndNotice{}  

\begin{abstract}
  Humans can understand and produce new utterances effortlessly,
  thanks to their compositional skills. Once a person
  learns the meaning of a new verb ``dax,'' he or she can immediately
  understand the meaning of ``dax twice'' or ``sing and dax.'' In this
  paper, we introduce the SCAN domain, consisting of a set of simple
  compositional navigation commands paired with the corresponding
  action sequences. We then test the zero-shot generalization
  capabilities of a variety of recurrent neural networks (RNNs)
  trained on SCAN with sequence-to-sequence methods. We find that RNNs
  can make successful zero-shot generalizations when the differences
  between training and test commands are small, so that they can apply
  ``mix-and-match'' strategies to solve the task. However, when
  generalization requires systematic compositional skills (as in the
  ``dax'' example above), RNNs fail spectacularly. We conclude with a
  proof-of-concept experiment in neural machine translation,
  suggesting that lack of systematicity might be partially responsible
  for neural networks' notorious training data thirst.
\end{abstract}

\input{introduction}

\input{world}

\input{models}

\input{experiments}

\input{discussion}
\subsubsection*{Acknowledgments}
We thank the reviewers, Joost Bastings, Kyunghyun Cho, Douwe Kiela, Germ\'an Kruszewski, Adam Liska, Tomas Mikolov, Kristina
Gulordava, Gemma Boleda, Michael Auli, Matt Botvinick, Sam Bowman, Jeff Dean, Jonas Gehring,
David Grangier, Angeliki Lazaridou, Gary Marcus, Jason Weston, the CommAI team and the audiences at the Facebook
Dialogue Summit, 2017 Paris Syntax and Semantics Colloquium and
CLIC-it 2017 for feedback and advice. The SCAN tasks are based on the
navigation tasks available at: \url{https://github.com/facebookresearch/CommAI-env}

\bibliography{marco_edited,library_clean}
\bibliographystyle{icml2018}

\clearpage
\input{supplementary_appendix}

\end{document}

%% file: introduction.tex
\section{Introduction}

Human language and thought are characterized by \emph{systematic
  compositionality}, the algebraic capacity to understand and produce
a potentially infinite number of novel combinations from known
components \citep{Chomsky:1957,Montague:1970a}. For example, if a
person knows the meaning and usage of words such as ``twice,''
``and,'' and ``again,'' once she learns a new verb such as ``to dax''
she can immediately understand or produce instructions such as ``dax
twice and then dax again.'' This type of compositionality is central
to the human ability to make strong generalizations from very limited
data \citep{Lake:etal:2016}.  In a set of influential and
controversial papers, Jerry Fodor and other researchers have argued
that neural networks are not plausible models of the mind because they
are associative devices that cannot capture systematic
compositionality
\citep[][a.o.]{Fodor:Pylyshyn:1988,Marcus1998,Fodor:Lepore:2002,Marcus2003,Calvo:Symons:2014}.

In the last few years, neural network research has made
astounding progress in practical domains where success depends on generalization. Perhaps most strikingly, \emph{end-to-end recurrent neural
  networks} currently dominate the state-of-the-art in machine
translation \citep{Bojar:etal:2016,Wu:etal:2016}. Since the overwhelming majority of sentences or even word sequences
in a language only occur once, even in a large corpus
\citep{Baroni:2009}, this points to strong generalization abilities.
Still, it is commonly observed that neural networks are extremely
sample inefficient, requiring very large training sets, which suggests
they may lack the same algebraic compositionality that humans exploit,
and they might only be sensitive to broad patterns over lots of
accumulated statistics \citep{Lake:etal:2016}.

In this paper, we introduce a grounded navigation environment where
the learner must translate commands given in a limited form of natural
language into a sequence of actions. This problem is naturally framed
as a sequence-to-sequence task, and, due to its simplicity, it is
ideal to study systematic generalization to novel examples in a
controlled setup. We thus use it to test a wide range of modern
recurrent network architectures in terms of their compositional
abilities. Our results suggest that standard recurrent seq2seq
architectures generalize very well when novel examples feature a
mixture of constructions that have been observed in training. However,
the models are catastrophically affected by \emph{systematic}
differences between training and test sentences, of the sort that
would be trivial for an agent equipped with an ``algebraic mind''
\citep{Marcus2003}.

%% file: world.tex
\section{The SCAN tasks}

We call our data set SCAN because it is a \textbf{S}implified version
of the \textbf{C}omm\textbf{A}I \textbf{N}avigation tasks
\citep[][]{Mikolov2015}.\footnote{SCAN available at:
  \url{https://github.com/brendenlake/SCAN}} For a learner, the goal
is to translate commands presented in simplified natural language into
a sequence of actions. Since each command is unambiguously associated
to a single action sequence, SCAN (unlike the original CommAI tasks)
can be straightforwardly treated as a supervised sequence-to-sequence
semantic parsing task
\citep{Dong:Lapata:2016,Jia:Liang:2016,Herzig:Berant:2017}, where the
input vocabulary is given by the set of words used in the commands,
and the output by the set of actions available to the learner.

\begin{figure*}[tb]
  \centering
  \begin{tabular}{lll}
    jump&$\Rightarrow$&JUMP\\
    jump left&$\Rightarrow$&LTURN JUMP\\
    jump around right&$\Rightarrow$&RTURN JUMP RTURN JUMP RTURN JUMP RTURN JUMP\\
    turn left twice&$\Rightarrow$&LTURN LTURN\\
    jump thrice&$\Rightarrow$&JUMP JUMP JUMP\\
    jump opposite left and walk thrice&$\Rightarrow$&LTURN LTURN JUMP WALK WALK WALK\\
    jump opposite left after walk around left&$\Rightarrow$&LTURN WALK LTURN WALK LTURN WALK LTURN WALK\\
    &&LTURN LTURN JUMP\\
  \end{tabular}
  \caption{Examples of SCAN commands (left) and the corresponding action sequences (right).}
  \label{fig:scan-examples}
\end{figure*}

Several examples from SCAN are presented in
Fig.~\ref{fig:scan-examples}.  Formally, SCAN consists of all the
commands generated by a phrase-structure grammar (presented in
Supplementary) and the corresponding sequence of actions, produced
according to a semantic interpretation function (see Supplementary).
Intuitively, the SCAN grammar licenses commands denoting primitive
actions such as JUMP (denoted by ``jump'';
Fig.~\ref{fig:scan-examples}), WALK (denoted by ``walk'') and LTURN
(denoted by ``turn left''). We will refer to these as \emph{primitive
  commands}.\footnote{Introducing the primitive turning actions LTURN
  and RTURN considerably simplifies the interpretation function,
  compared to capturing orientation by specifying arguments to the
  movement actions (e.g., JUMP[L], JUMP[R]).} It also accepts a set of
modifiers and conjunctions that compositionally build expressions
referring to action sequences. The ``left'' and ``right'' modifiers
take commands denoting undirected primitive actions as input and
return commands denoting their directed counterparts (``jump left'';
Fig.~\ref{fig:scan-examples}). The ``opposite'' modifier produces an
action sequence that turns the agent backward in the specified
direction before executing a target action (``jump opposite left''),
while ``around'' makes the agent execute the action at each step while
turning around in the specified direction (``jump around right'';
Fig.~\ref{fig:scan-examples}). The ``twice/thrice'' modifiers trigger
repetition of the command they take scope over, and ``and/after''
combine two action sequences. Although the SCAN examples in
Fig.~\ref{fig:scan-examples} focus on the ``jump''/JUMP primitive,
each instance of JUMP can be replaced with either WALK, RUN, or LOOK
to generate yet more commands.  Many more combinations are possible as
licensed by the grammar. The input vocabulary includes 13 words, the
output 6 actions.

The SCAN grammar, lacking recursion, generates a finite but large set
of unambiguous commands (20,910, to be precise). Commands can be
decoded compositionally by applying the corresponding interpretation
function. This means that, if it discovers the right interpretation
function, a learner can understand commands it has not seen during
training. For example, the learner might have only observed the
primitive ``jump'' command during training, but if it has learned the
meaning of ``after'', ``twice'' and ``around left'' from other verbs,
it should be able to decode, zero-shot, the complex command: ``jump
around left after jump twice''.

%% file: models.tex
\section{Models and setup} \label{sec_model}

\begin{figure}
  \centering
    \includegraphics[width=0.5\textwidth]{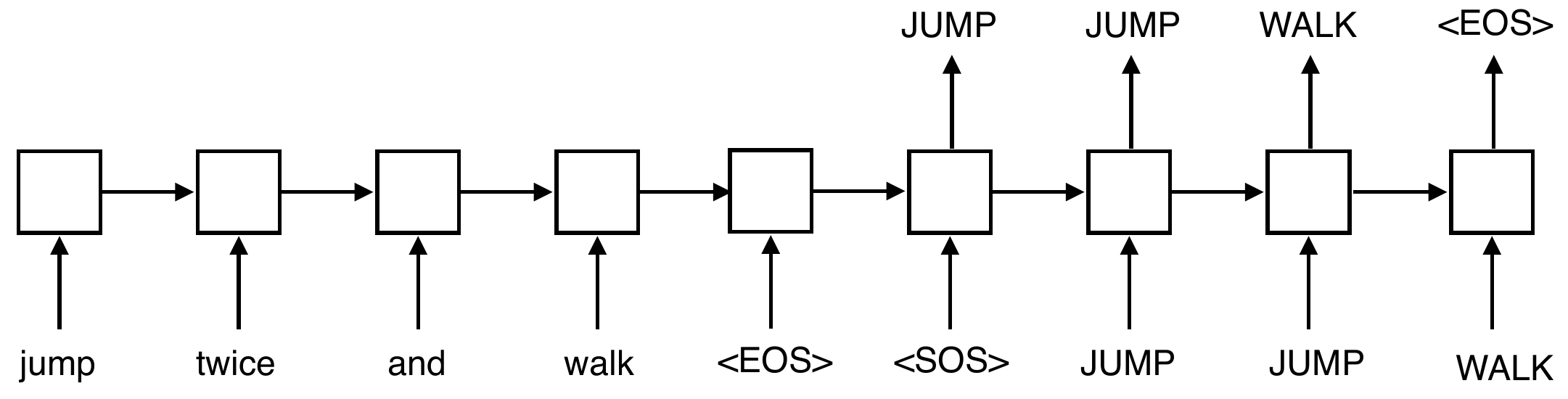}
    \caption{The seq2seq framework is applied to SCAN. The symbols $<$EOS$>$ and $<$SOS$>$ denote end-of-sentence and start-of-sentence, respectively. The encoder (left) ends with the first $<$EOS$>$ symbol, and the decoder (right) begins with $<$SOS$>$.}
     \label{fig_rnn_example}
\end{figure}

We approach SCAN through the successful sequence-to-sequence (seq2seq)
framework, in which two recurrent networks work together to learn a
mapping between input sequences and output sequences
\citep[e.g.,][]{Sutskever:etal:2014}.\footnote{Very recently,
  \emph{convolutional} seq2seq networks have reached comparable or
  superior performance in machine translation
  \citep{Gehring:etal:2017}. We will investigate them in future
  work.} %
Fig.~\ref{fig_rnn_example} illustrates the application of the seq2seq
approach to a SCAN example. First, a recurrent network encoder
receives the input sequence word-by-word, forming a low-dimensional
representation of the entire command. Second, the low-dimensional
representation is passed to a recurrent network decoder, which then
generates the output sequence action-by-action. The decoder's output
is compared with the ground truth, and the backpropagation algorithm
is used to update the parameters of both the encoder and
decoder. Note that although the encoder and decoder share the same
network structure (e.g., number of layers and hidden units), they do
not otherwise share weights/parameters with each other. More details
regarding the encoder-decoder RNN are provided in Supplementary.

Using the seq2seq framework, we tested a range of standard recurrent neural network models from the literature: simple recurrent networks \citep[SRNs;][]{Elman:1990}, long short-term memory networks \citep[LSTMs;][]{Hochreiter1997}, and gated recurrent units \citep[GRUs;][]{Chung:etal:2014}. Recurrent networks with attention have become increasingly popular in the last few years, and thus we also tested each network with and without an attentional mechanism, using the model from \citet[][]{Bahdanau:etal:2015} (see Supplementary for more details). Finally, to make the evaluations as systematic as possible, a large-scale hyperparameter search was conducted that varied the number of layers (1 or 2), the number of hidden units per layer (25, 50, 100, 200, or 400), and the amount of dropout (0, 0.1, 0.5; applied to recurrent layers and word embeddings). Varying these hyperparameters leads to 180 different network architectures, all of which were run on each experiment and replicated 5 times each with different random initializations.\footnote{A small number of runs (23/3600) did not complete, and thus not every network had 5 runs.}

In reporting the results, we focus on the \textbf{overall-best} architecture as determined by the extensive hyperparameter search. The winning architecture was a \textbf{2-layer LSTM with 200 hidden units per layer, no attention, and dropout applied at the 0.5 level}. Although the detailed analyses to follow focus on this particular model, the top-performing architecture for each experiment individually is also reported and analyzed.

Networks were trained with the following specifications. Training
consisted of 100,000 trials, each presenting an input/output sequence
and then updating the networks weights.\footnote{Note that, in all
  experiments, the number of \emph{distinct} training commands is well
  below 100k: we randomly sampled them with replacement to reach the
  target size} The ADAM optimization algorithm was used with
default parameters, including a learning rate of 0.001
\citep[][]{Kingma2014}. Gradients with a norm larger than 5.0 were
clipped. Finally, the decoder requires the previous step's output as
the next step's input, which was computed in two different
ways. During training, for half the time, the network's self-produced
outputs were passed back to the next step, and for the other half of
the time, the ground-truth outputs were passed back to the next step
\citep[teacher forcing;][]{Williams1989}. The networks were implemented in PyTorch and
based on a standard seq2seq
implementation.\footnote{\label{note_tutorial}The code we used is
  publicly available at the link:
  \url{http://pytorch.org/tutorials/intermediate/seq2seq_translation_tutorial.html}}

Training accuracy was above 99.5\% for the overall-best network in
each of the key experiments, and it was at least 95\% for the
top-performers in each experiment specifically.

%% file: experiments.tex
\section{Experiments}
In each of the following experiments, the recurrent networks are
trained on a large set of commands from the SCAN tasks to establish
background knowledge as outlined above. After training, the networks
are then evaluated on new commands designed to test generalization
beyond the background set in systematic, compositional ways. In
evaluating these new commands, the networks must make zero-shot
generalizations and produce the appropriate action sequence based
solely on extrapolation from the background training.

\subsection*{Experiment 1: Generalizing to a random subset of commands}

\label{sec:random_subset}
In this experiment, the SCAN tasks were randomly split into a training set (80\%) and a test set (20\%). The training set provides broad coverage of the task space, and the test set examines how networks can decompose and recombine commands from the training set. For instance, the network is asked to perform the new command, ``\textbf{jump opposite right after walk around right thrice},'' as a zero-shot generalization in the test set. Although the conjunction as a whole is novel, the parts are not: The training set features many examples of the parts in other contexts, e.g., ``\textbf{jump opposite right after} turn opposite right'' and ``jump right twice \textbf{after walk around right thrice}'' (both bold sub-strings appear 83 times in the training set). To succeed, the network needs to generalize by recombining pieces of existing commands to interpret new ones.

Overall, the networks were highly successful at generalization. The top-performing network for this experiment achieved 99.8\% correct on the test set (accuracy values here and below are averaged over the five training runs). The top-performing architecture was a LSTM with no attention, 2 layers of 200 hidden units, and no dropout. The best-overall network achieved 99.7\% correct. Interestingly, not every architecture was successful: Classic SRNs performed very poorly, and the best SRN achieved less than 1.5\% correct at test time (performance on the training set was equally low). However, attention-augmented SRNs learned the commands much better, achieving 59.7\% correct on average for the test set (with a range between 18.4\% and 94.0\% across SRN architectures). For LSTMs and GRUs, attention was instead not essential. Since the SCAN commands are never longer than 9 words,  attention is probably superfluous, at least in this simple setup, for gated architectures that generally exhibit a more robust long-distance behaviour than SRNs.

\begin{figure}[tb]
  	\centering
    \includegraphics[width=0.49\textwidth]{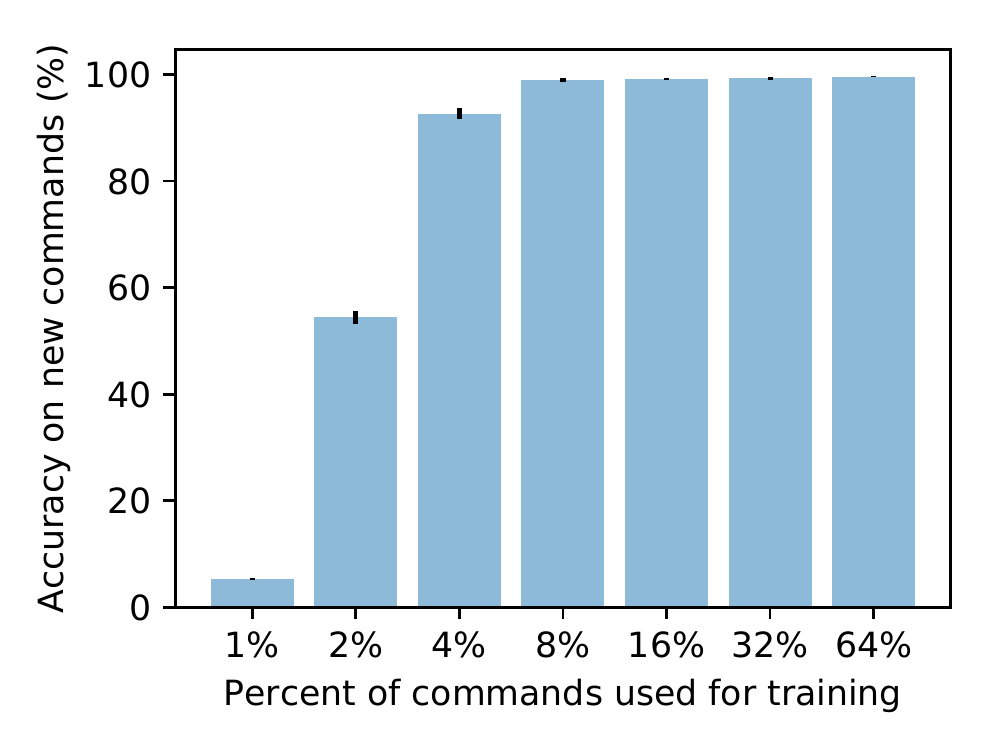}
    \caption{Zero-shot generalization after training on a random
      subset of the SCAN tasks. The overall-best network was trained
      on varying proportions of the set of distinct tasks (x-axis) and
      generalization was measured on new tasks (y-axis). Each bar
      shows the mean over 5 training runs with corresponding $\pm$1
      SEM.}
     \label{fig_sample_complexity}
\end{figure}


As indicated above, the main split was quite generous, providing 80\%
of the commands at training time for a total of over 16,700 distinct
examples (with strong combinatorial coverage).  We next re-trained the
best-overall network with varying numbers of \emph{distinct} examples
(the actual number of training \emph{presentations} was kept constant at
100K). The results are shown in Fig.~\ref{fig_sample_complexity}. With
1\% of the commands shown during training (about 210 examples), the
network performs poorly at about 5\% correct. With 2\% coverage,
performance improves to about 54\% correct on the test set. By 4\%
coverage, performance is about 93\% correct. Our results show that not
only can networks generalize to random subsets of the tasks, they can
do so from relatively sparse coverage of the compositional command
space.  This is well in line with the success of seq2seq architectures
in machine translation, where most test sentences are likely never
encountered in training. Still, even with this sparser coverage,
differences between training and test instances are not
dramatic. Let us for example consider the set of all commands without a
conjunction (e.g., ``walk around thrice'', ``run'', ``jump opposite
left twice''). All the commands of this sort that occur in the test
set of the 2\% training coverage split (either as components of a
conjunction or by themselves) \emph{also} occur in the corresponding
training set, with an average of 8 occurrences. Even for the 1\%
split, there is only one conjunction-less test command that does not
also occur in the training split, and the frequency of occurrence of
such commands in the training set is at a non-negligible average value
of 4 times.

\input{experiment2}

\input{experiment3}

\input{experiment4}

%% file: experiment2.tex
\subsection*{Experiment 2:  Generalizing to commands demanding longer action sequences}



We study next a more \emph{systematic} form of generalization, where
models must bootstrap to commands requiring longer action sequences
than those seen in training.\footnote{We focus on action sequence
  length rather than command length since the former exhibits more
  variance (1-48 vs.~1-9). The longest commands (9 words) are given by the
  conjunction of two directed primitives both modified twice, e.g.: ``jump around left twice and run opposite right
  thrice.'' On the other hand, a relatively short command such as
  ``jump around left thrice'' demands 24 actions.} Now the training
set contains all 16,990 commands requiring sequences of up to 22
actions, whereas the test set includes all remaining commands (3,920,
requiring action sequences of lengths from 24 to 48). Under this
split, for example, at test time the network must execute the command
``jump around left twice and walk opposite right thrice'', requiring a
sequence of 25 actions. Although all the elements used in the command
have been observed during training, the network has never been asked
to produce a sequence of this length, nor it has ever seen an ``around
* twice'' command conjoined with an ``opposite * thrice'' command
(although it did observe both components conjoined with others). Thus,
it must productively generalize familiar verbs, modifiers and
conjunctions to generate longer action sequences. This is a
fair task for a system that is correctly translating the input
commands. If you know how to ``walk around,'' how to ``jump,'' and the
function of the ``and'' conjunction, you will be immediately able to
``walk around and jump,'' even if you have never performed an action
sequence of that length.

This test turns out to be very challenging for all models. The best
result (20.8\% on average, again over 5 runs) is achieved by a GRU with attention, one
50-dimensional hidden layer, and dropout 0.5. Interestingly, this is a model with considerably
less capacity than the best for the random-split setup, but it uses attention, which might help, to a limited degree, to generalize to longer action sequences. The overall-best model achieves 13.8\% accuracy.

\begin{figure}[tb]
  	\centering
        \includegraphics[width=0.49\textwidth]{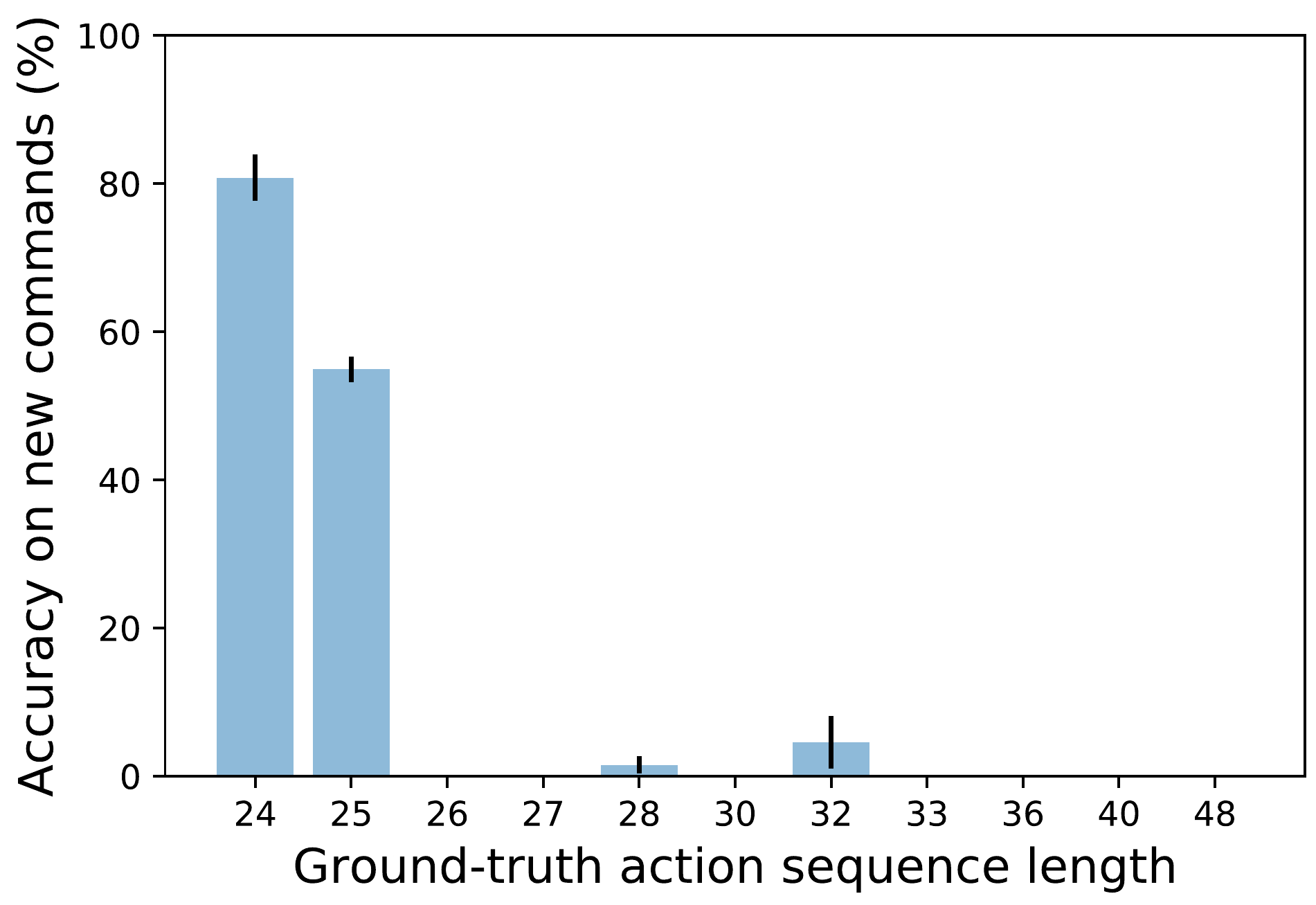}
        \includegraphics[width=0.49\textwidth]{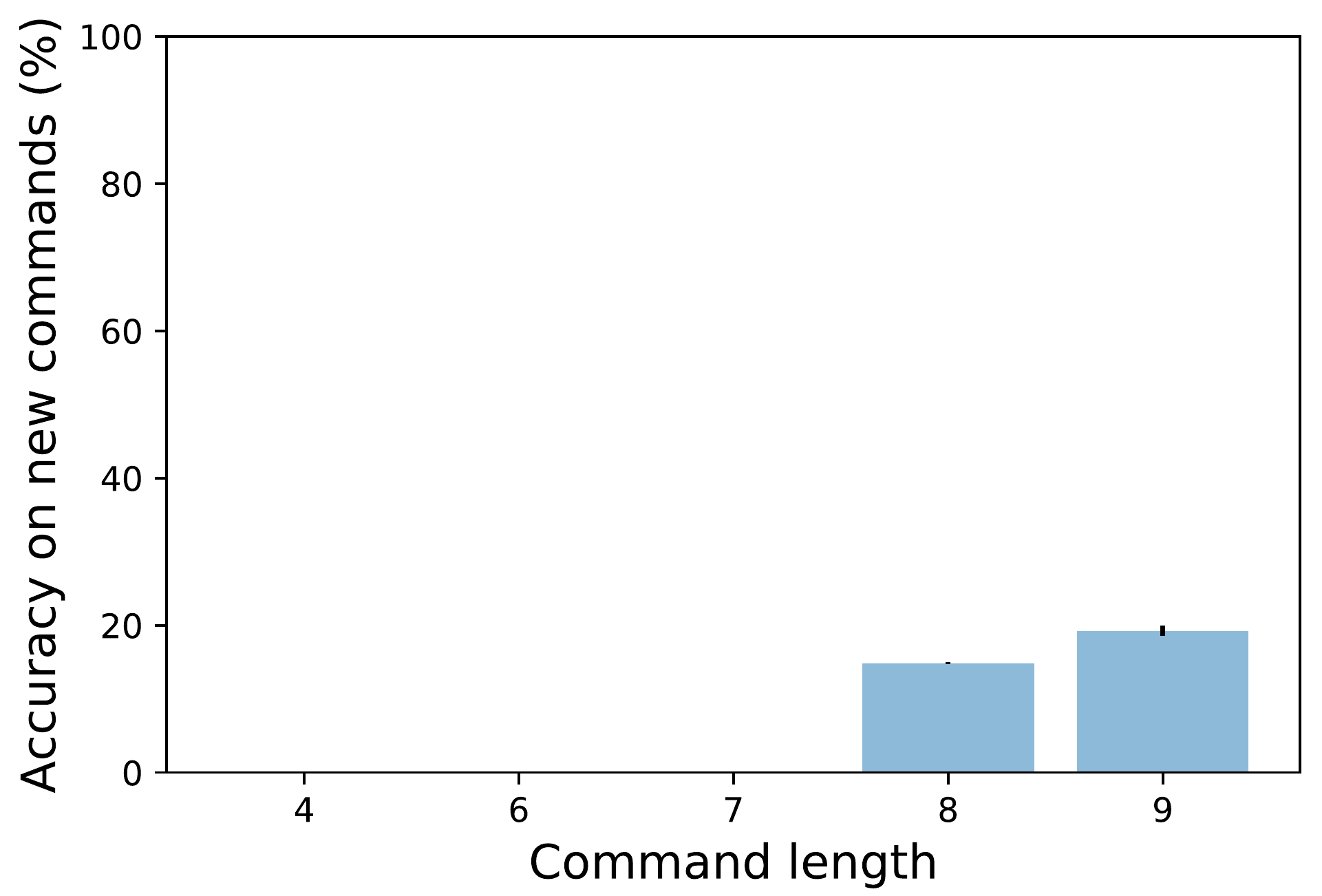}
    \caption{Zero-shot generalization to commands with action sequence lengths not seen in training. Top: accuracy distribution by action sequence length. Bottom: accuracy distribution by command length (only lengths attested in the test set shown, in both cases). Bars show means over 5 runs of overall-best model with  $\pm$1 SEM.}
     \label{fig_acc_by_length}
\end{figure}

Fig.~\ref{fig_acc_by_length} (top) shows partial success is almost entirely explained by
generalization to the shortest action sequence lengths in the test
set. Although we might not expect even humans to be able to generalize to
very long action sequences, the sharp drop between extrapolating to 25
and 26 actions is striking. %
%
The bottom panel of Fig.~\ref{fig_acc_by_length} shows accuracy in the test set
organized by command length (in word tokens). %
The model only gets right
some of the \emph{longest} commands (8 or 9 tokens). In the training set, the longest action sequences
($\geq$20) are invariably associated to commands containing 8 or 9
tokens. Thus, the model is correctly generalizing only
in those cases that are most similar to training instances.  

Finally, we performed two additional analyses to better understand the source of the errors. 
First, we examined the greedy decoder for search-related errors. We confirmed that, for almost every error, the network preferred its self-generated output sequence to the target output sequence (as measured by log-likelihood). Thus, the errors were not due to search failures in the decoder.\footnote{For both the overall best model and the best model in this experiment, on average over runs, less than one test command (of thousands) could be attributed to a search failure.}
Second, we studied whether the difficulty with long sequences can be
mitigated if the proper length was provided by an oracle at evaluation
time.\footnote{Any attempt from the decoder to terminate the action
  sequence with an $<$EOS$>$ was ignored (and the second strongest
  action was chosen) until a sequence with proper length was
  produced.} If this difficulty is a relatively straightforward issue
of the decoder terminating too early, then this should provide an
(unrealistic) fix. If this difficulty is symptomatic of deeper
problems with generalization, then this change will have only a small
effect. With the oracle, the overall-best network performance improved
from 13.8\% to 23.6\% correct, which was notable but insufficient to
master the long sequences. The top-performing model showed a more
substantial improvement (20.8\% to 60.2\%). Although improved, the
networks were far from perfect and still exhibited difficulties
with long sequences of output actions (again, even for the top model,
there was a strong effect of action sequence length, with average
accuracy ranging from 95.76\% for commands requiring 24 actions to 22.8\%
for commands requiring 48 actions).

%% file: experiment3.tex
\subsection*{Experiment 3: Generalizing composition across primitive commands}

Our next test is closest to the ``dax'' thought experiment presented in the introduction. In the training phase, the model is exposed to the
primitive command only denoting a certain basic action (e.g.,
``jump''). The model is also exposed to all primitive and composed
commands for all other actions (e.g., ``run'', ``run twice'',
``walk'', ``walk opposite left and run twice'', etc.). At test time,
the model has to execute all composed commands for the action that
it only saw in the primitive context (e.g., ``jump twice'', ``jump
opposite left and run twice'', etc.). According to the classic thought
experiments of Fodor and colleagues, this should be easy: if you know
the meaning of ``run'', ``jump'' and ``run twice'', you should also
understand what ``jump twice'' means.

We run two variants of the experiment generalizing from ``turn left''
and ``jump'', respectively. Since ``turn right'' is distributionally
identical to ``turn left'' (in the sense that it occurs in exactly the
same composed commands) and ``walk'', ``run'' and ``look'' are
distributionally identical to ``jump'', it is redundant to test all
commands. Moreover, to ensure the networks were highly familiar with
the target primitive command (``jump'' or ``turn left''), the latter was
over-represented in training such that roughly 10\% of all training
presentations were of the command.\footnote{Without over-sampling, performance was consistently worse than what we report.}

We obtain strikingly different results for ``turn left'' and
``jump''. For ``turn left'', many models generalize very well to
composed commands. The best performance is achieved by a GRU network with attention, one layer with 100 hidden units, and dropout of 0.1
(90.3\% accuracy). The overall-best model achieved 90.0\%
accuracy. On the other hand, for ``jump,'' models are almost completely
incapable to generalize to composed commands. The best performance was 1.2\% accuracy (LSTM, attention, one layer, 100 hidden units, dropout 0.1). The overall-best
model reached 0.08\% accuracy. As in Experiment 2, the errors were not due to search failures in the decoder.

\begin{table*}[tb]
  \centering
  \caption{Nearest training commands for representative commands, with the respective cosines. Here, ``jump'' was trained in isolation
    while ``run'' was trained compositionally. Italics mark
    low similarities (cosine $<$0.2).}
  \label{tab:nns}
  \begin{tabular}{ll|ll|ll|ll}
    \multicolumn{2}{c|}{run}&\multicolumn{2}{c|}{jump}&\multicolumn{2}{c|}{run twice}&\multicolumn{2}{c}{jump twice}\\
    \hline
    look                &.73&\emph{run}                              &\emph{.15}&look twice                              &.72&\emph{walk and walk}&                \emph{.19}\\ \hline
    walk                &.65&\emph{walk}            &\emph{.13}&run twice and&.65&\emph{run and walk}&        \emph{.16}\\ 
                    &&            &&look opposite right thrice&&&        \\ \hline
    walk after run      &.55&\emph{turn right}&\emph{.12}&run twice and  &.64&\emph{walk opposite right}& \emph{.12}\\ 
                        &   &        &          &run right twice&   &\emph{and walk}                  & \\ \hline
    run thrice&.50&\emph{look right twice}&\emph{.09}&run twice and            &.63&\emph{look right and walk}&\emph{.12}\\
    after run &   &\emph{after walk twice}&          &look opposite right twice&   &                          &          \\ \hline 
    run twice&.49&\emph{turn right}      &\emph{.09}&walk twice and run twice       &.63&\emph{walk right and walk}&       \emph{.11}\\
     after run &&\emph{after turn right}      &&       &&&       \\ 
  \end{tabular}
\end{table*}

In the case of ``turn left'', although models are only exposed to the
primitive command during training, they will see the action it denotes
(LTURN) many times, as it is used to accomplish many directed
actions. For example, a training item is: ``walk left and jump
left'', with ground-truth interpretation: LTURN WALK LTURN
JUMP. Apparently, seeing \emph{action sequences} containing LTURN
suffices for the model to understand composed \emph{commands} with
``turn left'', probably because the model receives direct evidence
about how LTURN is used in context. On the other hand, the
action denoted by ``jump'' (JUMP) only occurs with this primitive
command in training, and the model does not generalize from this
minimal context to new composed ones.

We now take a closer look at the results, focusing on the
median-performance run of the overall-best model (as the most
representative run of this model). We observe that even in the
successful ``turn left'' case model errors are surprising. One
would expect such errors to be randomly distributed, or perhaps to
pertain to the longest commands or action sequences. Instead, all 45
errors made by the model are conjunctions where one of the components
is simple ``turn left'' (22 cases) or ``turn left thrice'' (23
cases). This is particularly striking because the network produced the
correct mapping for ``turn left'' during training, as well as for
``turn left thrice'' at test time, and it gets many more conjunctions
right (ironically, including ``turn left thrice and turn left'',
``turn left thrice after turn left'' etc.). We conclude that, even
when the network has apparently learned systematic composition almost
perfectly, it got at it in a very counter-intuitive way. It's hard to
conceive of someone who understood the meaning of ``turn left'', and
``jump right and turn left twice'' (which the network gets right), but
not that of ``jump right and turn left'' (one of the examples the
network missed). In the ``jump'' experiment, the network could only
correctly decode two composite cases, both starting with the execution
of primitive ``jump'', conjoined with a different action: ``jump and
run opposite right'', ``jump and walk around left thrice''.

It is instructive to look at the representations that the network
induced for various commands in the latter experiment. Table
\ref{tab:nns} reports the 5 nearest neighbours for a sample of
commands. Command similarity is measured by the cosine between the
final encoder hidden state vectors, and computed with respect to all
commands present in the training set.  ``Run'' is provided as an
example primitive command for which the model has been exposed to the
full composed paradigm in training. As one would expect, ``run'' is
close to the other primitive commands (``look'', ``walk''), as well as
to short conjoined commands that contain primitive ``run'' as one of
the conjuncts (we observe a similar pattern for the non-degenerate ``jump''
representation induced in Experiment 1). Instead, since ``jump'' had a different
training distribution than the other primitive commands, the model
does not capture its similarity to them, as shown by the very low
cosines of its nearest commands. Since it fails to establish a
link to other basic commands, the model does not generalize
modifier application from them to ``jump''. Although ``run twice'' is
similar to (conjunctions of) other primitive tasks composed with
``twice'', ``jump twice'' is isolated in representational space, and
its (far) nearest neighbours look arbitrary.

We tested here systematicity in its purest form: the model was only
exposed to ``jump'' in isolation, and asked to bootstrap to its
compositional paradigm based on the behaviour of other primitive
commands such as ``walk'', ``look'' and ``run''. Although we suspect
humans would not have problems with this setup, it arguably is
too opaque for a computational model, which could lack evidence for
``jumping'' being the same sort of action as ``walking''. Suppose we
give the network \emph{some} evidence that ``jumping'' composes like
``walking'' by showing a few composed ``jump'' command during
training. Is the network then able to generalize to the full composed
paradigm?

\begin{figure}[tb]
  \centering
    \includegraphics[width=0.49\textwidth]{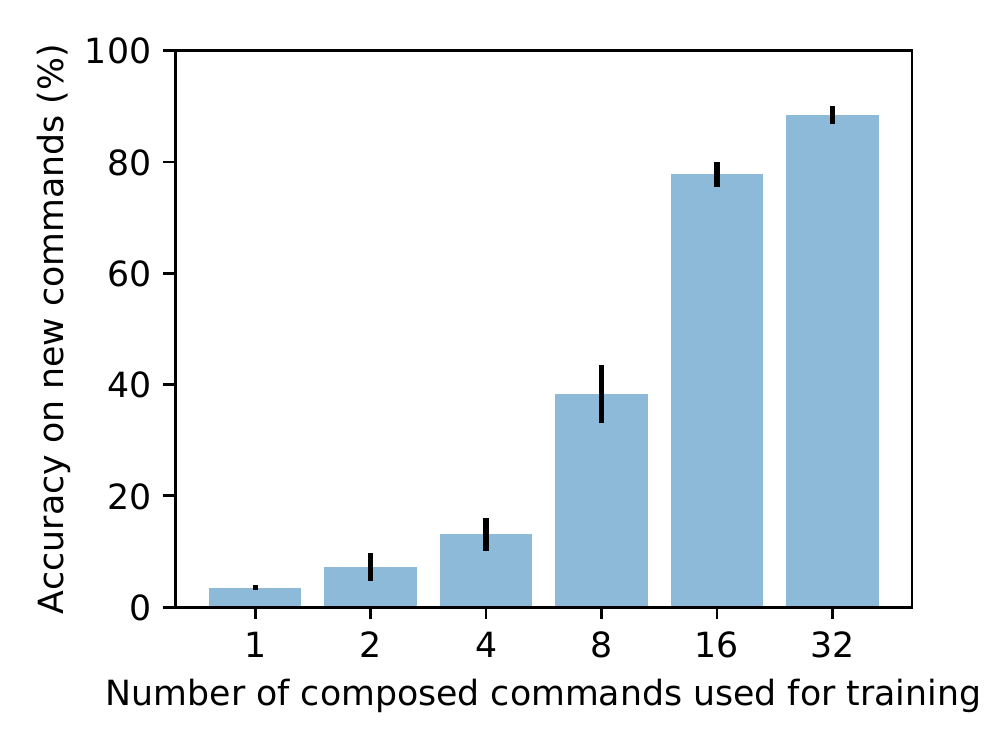}
    \caption{Zero-shot generalization after adding the primitive ``jump'' and some compositional ``jump'' commands. The model that performed best in generalizing from primitive ``jump'' only was re-trained with different numbers of composed ``jump'' commands (x-axis) in the training set, and generalization was measured on new composed ``jump'' commands (y-axis). Each bar shows the mean over 5 runs with varying training commands along with the corresponding $\pm$1 SEM.}
     \label{fig:jump_compositions}
\end{figure}

This question is answered in Figure \ref{fig:jump_compositions}.  We
present here results for the best model in the
``jump''-generalization task, which was noticeably better in the
present setup than the overall-best model. Again, the new primitive
command (and its compositions) were over-sampled during training to
make up 10\% of all presentations.  Here, even when shown 8 different
composed commands with ``jump'' at training time, the network only weakly
generalizes to other composed commands (38.3\% correct). Significant generalization
(still far from systematic) shows up when the training set contains 16
and especially 32 distinct composed commands (77.8\% and 88.4\%,
respectively).  We conclude that the network is not failing to
generalize simply because, in the original setup, it had little evidence
that ``jump'' should behave like the other commands. On the other
hand, the runs with more composed examples confirm that, as we found
in Experiment 1, the network does display powerful generalization
abilities. Simply, they do not conform to the ``all-or-nothing''
rule-based behaviour we would expect from a systematically
compositional device--and, as a consequence, they require more
positive examples to emerge.


%% file: experiment4.tex
\subsection*{Experiment 4: Compositionality in machine translation}
Our final experiment is a proof-of-concept that our findings
are more broadly applicable; that is, the limitations of recurrent 
networks with regards to systematic compositionality extend beyond SCAN to
other sequence-to-sequence problems such as machine translation. First, we trained our standard seq2seq code on short ($\le 9$ words) English-French sentence pairs that begin with English phrases such as ``I am,'' ``he is,'' ``they are,'' and their contractions (randomly split with 10,000 for training and 1180 for testing).\footref{note_tutorial}
An informal hyperparameter search led us to pick a LSTM with attention, 2 layers of 400 hidden units, and 0.05 dropout. With these hyperparameters and the same training procedure used for the SCAN tasks (Section \ref{sec_model}), the network reached a respectable 28.6 BLEU test score after 100,000 steps.

Second, to examine compositionality with the introduction of a new
word, we trained a fresh network after adding 1,000 repetitions of the
sentence ``I am daxy'' (fr.~``je suis daxiste'') to the training data
(the BLEU score on the original test set dropped less than 1
point).\footnote{Results do not change if, instead
  of repeating ``I am daxy'' 1,000 times, we insert it 100 times; with just 1 or
  10 occurrences of this sentence in the training data, we get 0/8
  translations right.} %
We tested this network by embedding ``daxy'' into the
following constructions: ``you are daxy'' (``tu es daxiste''), ``he is
daxy'' (``il est daxiste''), ``I am not daxy'' (``je ne suis pas
daxiste''), ``you are not daxy'' (``tu n'es pas daxiste''), ``he is
not daxy'' (``il n'est pas daxiste''), ``I am very daxy'' (``je suis
tr\`es daxiste''), ``you are very daxy'' (``tu es tr\`es daxiste''),
``he is very daxy'' (``il est tr\`es daxiste''). During training, the
model saw these constructions occurring with 22 distinct predicates on
average (limiting the counts to perfect matches, excluding, e.g.,
``you are not very X''). Still, the model could only get one of the 8
translations right (that of ``he is daxy''). For comparison, for the
adjective ``tired'', which occurred in 80 different constructions in
the training corpus, our network had 8/8 accuracy when testing on the
same constructions as for ``daxy'' (only one of which also occurred
with ``tired'' in the training set). Although this is a small-scale
machine translation problem, our preliminary result suggests that
models will similarly struggle with systematic compositionality in
larger data sets, when adding a new word to their vocabulary, in ways
that people clearly do not.

%% file: discussion.tex
\section{Discussion}

In the thirty years since the inception of the systematicity debate,
many have tested the ability of neural networks to solve tasks
requiring compositional generalization, with mixed results
\citep[e.g.,][]{Christiansen:Chater:1994,Marcus1998,Phillips:1998,Chang:2002,vanderVelde:etal:2004,Botvinick:Plaut:2006,Wong:Wang:2007,Bowers:etal:2009,Botvinick:plaut:2009,Brakel:Frank:2009,Frank:etal:2009,Frank:2014,Bowman2015Tree}. However,
to the best of our knowledge, ours is the first study testing
systematicity in modern seq2seq models, and our results confirm the
mixed picture. On the one hand, Experiment 1 and the ``turn left''
results in Experiment 3 show how standard recurrent models can reach
very high zero-shot accuracy from relatively few training examples. We
would like to stress that this is an important positive result,
showing in controlled experiments that seq2seq models can make 
powerful zero-shot generalizations. Indeed, an interesting direction for future work is to
understand what are, precisely, the generalization mechanisms that
subtend the networks' success in these experiments. 
After all, human language does have plenty of generalization patterns that are not
easily accounted for by algebraic compositionality
\citep[see, e.g.,][]{Goldberg:2005}.

On the other hand, the same networks fail spectacularly when the
link between training and testing data is dependent on the ability to
extract \emph{systematic} rules. This can be seen as a trivial
confirmation of the basic principle of statistical machine learning that
your training and test data should come from the same
distribution. But our results also point to an important difference
in how humans and current seq2seq models generalize, since there is no doubt that human learners can
generalize to unseen data when such data are governed by rules that
they have learned before. Importantly, the training data of experiments 2 and 3
provide enough evidence to learn composition rules affording the
correct generalizations. In Experiment 2, the training data contain
examples of all modifiers and connectives that are needed at test time
for producing longer action sequences. In Experiment 3, the usage of
modifiers and connectives is illustrated at training time by their
application to many combinations of different primitive commands, and, at test time, the network
should apply them to a new command it encountered in isolation during
training.

We thus believe that the fundamental component that current models are
missing is the ability to extract systematic rules from the training
data. A model that can abstract away from surface statistical patterns and
operate in ``rule space'' should extract rules such as: translate(\emph{x} and \emph{y}) = translate(\emph{x}) translate(\emph{y}); translate(\emph{x} twice) = translate(\emph{x}) translate(\emph{x}). Then, if the
meaning of a new command (translate(``jump'')) is learned at
training time, and acts as a variable that rules can be applied to, no
further learning is needed at test time. When represented in this more
abstract way, the training and test distributions are quite similar,
even if they differ in terms of shallower statistics such as word
frequency.

How can we encourage seq2seq models to extract rules from
data rather than exploiting shallower pattern recognition mechanisms?
We think there are several, non-mutually exclusive avenues to be explored.

First, in
a ``learning-to-learn'' approach
\citep[][a.o.]{Thrun:Pratt:1997,Risi:etal:2009,Finn:etal:2017}, a network
can be exposed to a number of different learning environments regulated
by similar rules. An objective function requiring successful
generalization to new environments might encourage learners to discover the
shared general rules.

Another promising approach is to add more structure to the neural networks.
Taking inspiration from recent neural program induction and modular network models \citep[e.g.,][]{Reed:deFreitas:2016,Hu:etal:2017,Johnson2017}, we could endow RNNs with a set of manually-encoded or
(ideally) learned functions for interpreting individual modifiers, connectives, and primitives.
The job of the RNN would be to learn how to apply and compose these functions as appropriate
for interpreting a command. Similarly, differentiable stacks, tapes, or random-access memory \citep[e.g.,][]{Joulin:Mikolov:2015,Graves:etal:2016} could equip seq2seq models with quasi-discrete memory structures, enabling separate storage of variables, which in turn might encourage abstract rule learning \citep[see ][for a memory-augmented seq2seq model]{Feng:etal:2017}.

Other solutions, such as \emph{ad-hoc} copying mechanisms or special
ways to initialize the embeddings of novel words, might help to solve
the SCAN tasks specifically. But they are unlikely to help with more
general seq2seq problems. It remains to be seen, of course, if 
any of our proposed approaches offer a truly general solution. 
Nonetheless, we see all of the suggestions as directions worth pursuing, perhaps
simultaneously and in complementary ways, with the goal of achieving
human-like systematicity on SCAN and beyond.

Given the astounding successes of seq2seq models in
challenging tasks such as machine translation, one might argue that
failure to generalize by systematic composition indicates that neural
networks are poor models of some aspects of human cognition, but it is of little
practical import. However, systematicity is an extremely efficient way to
generalize. Once a person learns the new English adjective ``daxy'', he or she can
immediately produce and understand an infinity of sentences containing
it. The SCAN experiments and a proof-of-concept machine
translation experiment (Experiment 4) suggest that this ability is still beyond
the grasp of state-of-the-art neural networks, likely contributing to 
their striking need for very large training sets. These results
give us hope that neural networks capable of systematic compositionality could greatly
benefit machine translation, language modeling, and other applications.

%% file: supplementary_appendix.tex
\section*{Supplementary materials}

\subsection*{SCAN grammar and interpretation function}

The phrase-structure grammar generating all SCAN commands is presented
in Figure \ref{fig:scan-psg}. The corresponding interpretation
functions is in Figure \ref{fig:scan-int}.

\begin{figure*}[b]
  \centering
  \begin{tabular}{lll}
    C $\to$ S and S  &V $\to$ D[1] opposite D[2]&D $\to$ turn left\\
    C $\to$ S after S&V $\to$ D[1] around D[2]  &D $\to$ turn right\\
    C $\to$ S        &V $\to$ D                 &U $\to$ walk\\
    S $\to$ V twice  &V $\to$ U                 &U $\to$ look\\
    S $\to$ V thrice &D $\to$ U left            &U $\to$ run\\
    S $\to$ V        &D $\to$ U right           &U $\to$ jump\\
  \end{tabular}
  \caption{Phrase-structure grammar generating SCAN commands. We use indexing notation to allow infixing: D[i] is to be read as the i-th element directly dominated by category D.}
  \label{fig:scan-psg}
\end{figure*}

\begin{figure*}[b]
  \centering
  \begin{tabular}{ll}
    $\llbracket$walk $\rrbracket$ = WALK                              &$\llbracket u$ opposite left$\rrbracket$ = $\llbracket$turn opposite left$\rrbracket$ $\llbracket u \rrbracket$\\
    $\llbracket$look$\rrbracket$ = LOOK                               &$\llbracket u$ opposite right$\rrbracket$ = $\llbracket$turn opposite right$\rrbracket$ $\llbracket u \rrbracket$\\
    $\llbracket$run$\rrbracket$ = RUN                                 &$\llbracket$turn around left$\rrbracket$ = LTURN LTURN LTURN LTURN\\                               
    $\llbracket$jump$\rrbracket$ = JUMP                               &$\llbracket$turn around right$\rrbracket$ = RTURN RTURN RTURN RTURN\\
    $\llbracket$turn left$\rrbracket$ = LTURN                         &$\llbracket u$ around left$\rrbracket$ = LTURN $\llbracket u \rrbracket$ LTURN $\llbracket u\rrbracket$ LTURN $\llbracket u\rrbracket$  LTURN $\llbracket u\rrbracket$\\
    $\llbracket$turn right$\rrbracket$ = RTURN                        &$\llbracket u$ around right$\rrbracket$ = RTURN $\llbracket u\rrbracket$ RTURN $\llbracket u\rrbracket$ RTURN $\llbracket u\rrbracket$  RTURN $\llbracket u\rrbracket$\\
    $\llbracket u$ left$\rrbracket$ = LTURN $\llbracket u \rrbracket$ &$\llbracket x$ twice$\rrbracket$ = $\llbracket x\rrbracket\ \llbracket x\rrbracket$\\
    $\llbracket u$ right$\rrbracket$ = RTURN $\llbracket u \rrbracket$&$\llbracket x$ thrice$\rrbracket$ = $\llbracket x \rrbracket\ \llbracket x\rrbracket\ \llbracket x\rrbracket$\\
    $\llbracket$turn opposite left$\rrbracket$ = LTURN LTURN          &$\llbracket x_1$ and $x_2\rrbracket$ = $\llbracket x_1\rrbracket \ \llbracket x_2\rrbracket$\\
    $\llbracket$turn opposite right$\rrbracket$ = RTURN RTURN         &$\llbracket x_1$ after $x_2\rrbracket$ = $\llbracket x_2 \rrbracket\ \llbracket x_1\rrbracket$\\
  \end{tabular}
  \caption{Double brackets ($\llbracket\rrbracket$) denote
    the interpretation function translating SCAN's linguistic commands
    into sequences of actions (denoted by uppercase strings). Symbols $x$ and $u$ denote variables, the latter limited to words in the set \{walk, look, run, jump\}. The linear order of actions denotes their temporal sequence.}
  \label{fig:scan-int}
\end{figure*}

\subsection*{Standard Encoder-Decoder RNN}
We describe the encoder-decoder framework, borrowing from the description in \citet{Bahdanau:etal:2015}. The encoder receives a natural language command as a sequence of $T$ words. The words are transformed into a sequence of vectors, $\{w_1, \dots, w_T\}$, which are learned embeddings with the same number of dimensions as the hidden layer. A recurrent neural network (RNN) processes each word
\[h_t = f_E(h_{t-1}, w_{t}),\]
where $h_t$ is the encoder hidden state. The final hidden state $h_T$ (which may include multiple layers for multi-layer RNNs) is passed to the RNN decoder as hidden state $g_0$ (see seq2seq diagram in the main article). Then, the RNN decoder must generate a sequence of output actions $a_1,\dots,a_R$. To do so, it computes
\[g_t = f_D(g_{t-1},a_{t-1}),\]
where $g_t$ is the decoder hidden state and $a_{t-1}$ is the (embedded) output action from the previous time step. Last, the hidden state $g_t$ is mapped to a softmax to select the next action $a_t$ from all possible actions.

\subsection*{Attention Encoder-Decoder RNN}
For the encoder-decoder with attention, the encoder is identical to the one described above. Unlike the standard decoder that can only see $h_T$, the attention decoder can access all of the encoder hidden states, $h_1,\dots,h_T$ (in this case, only the last layer if multi-layer). At each step $i$, a context vector $c_i$ is computed as a weighted sum of the encoder hidden states
\[c_i = \sum_{t=1}^T \alpha_{it} h_t.\]
The weights $\alpha_{it}$ are computed using a softmax function
\[\alpha_{it} = \exp(e_{it})/\sum_{j=1}^T\exp(e_{ij}),\]
where $e_{it} = v_a^{\top} \tanh(W_a g_{i-1} + U_a h_t)$ is an alignment model that computes the similarity between the previous decoder hidden state $g_{i-1}$ and an encoder hidden state $h_t$ (for the other variables, $v_a$, $W_a$, and $U_a$ are learnable parameters) \citep{Bahdanau:etal:2015}. This context vector $c_i$ is then passed as input to the decoder RNN at each step with the function
\[g_i = f_D(g_{i-1},a_{i-1},c_i),\]
which also starts with hidden state $g_0 = h_T$, as in the standard decoder. Last, the hidden state $g_i$ is concatenated with $c_i$ and mapped to a softmax to select new action $a_i$.